\documentclass[11pt,a4paper]{article}
\usepackage[dvipsnames]{xcolor}
\usepackage[hyperref]{acl2018}
\usepackage{times}
\usepackage{latexsym}
\usepackage[utf8]{inputenc}
\usepackage{graphicx}
\usepackage{multirow}
\usepackage{booktabs}
\usepackage{array}
\usepackage{subfigure}
\usepackage{tabularx}
\usepackage{amsmath}
\usepackage{latexsym}
\usepackage{graphicx}
\usepackage{amsmath}
\usepackage{enumitem}
\usepackage{amsfonts}
\usepackage{boxedminipage}
\usepackage{color}
\usepackage{bm}
\usepackage{amssymb}
\usepackage{times}
\usepackage{bbold}
\usepackage{array}
\usepackage{amssymb}
\usepackage{appendix}
\usepackage{sidecap}
\usepackage{mathtools}
\usepackage{wrapfig}
\usepackage{float}
\usepackage{tikz}
\usepackage[printwatermark]{xwatermark}

\aclfinalcopy
\def\aclpaperid{1615} 
\title{
Modeling Naive Psychology of Characters in Simple Commonsense Stories
}

\author{
  Hannah Rashkin$^\dagger$,
  Antoine Bosselut$^\dagger$,
  Maarten Sap$^\dagger$,
  Kevin Knight$^\ddagger$ \and Yejin Choi$^{\dagger\mathsection}$ \\
    $^\dagger$Paul G. Allen School of Computer Science \& Engineering, University of Washington \\
    $^\mathsection$Allen Institute for Artificial Intelligence\\
  {\tt \{hrashkin,msap,antoineb,yejin\}@cs.washington.edu}\\
$^\ddagger$ Information Sciences Institute \& Computer Science, University of Southern California\\
  {\tt knight@isi.edu }
  }

\begin{document}

\maketitle

\begin{abstract}
Understanding a narrative requires reading between the lines and reasoning about the unspoken but obvious implications about events and people's mental states --- a capability that is trivial for humans but 
remarkably hard for machines.  
%
To facilitate research addressing this challenge, 
we introduce a new annotation framework to explain naive psychology of story characters 
as fully-specified chains of mental states with respect to  \emph{motivations} and \emph{emotional reactions}. Our work presents a new large-scale dataset with rich low-level annotations and establishes baseline performance on several new tasks, 
suggesting avenues for future research.

\end{abstract}

\section{Introduction}
\label{sec:intro}

Understanding a story requires reasoning about the causal links between the events in the story and the mental states of the characters, even when those relationships are not explicitly stated.
As shown by the commonsense story cloze 
shared task \cite{SharedTask}, this reasoning is remarkably hard for both statistical and neural 
machine readers -- despite being trivial for humans. 
This stark performance gap between humans and machines is not surprising as most powerful language models have been designed to effectively learn local fluency patterns. Consequently, they generally lack the ability to abstract away from surface patterns in text to model more complex implied dynamics, such as intuiting characters' mental states or predicting their plausible next actions. 

\begin{figure}\label{fig:story_motivation}
\begin{centering}
\vspace*{-2mm}
\includegraphics[width=.95\columnwidth, page=1]{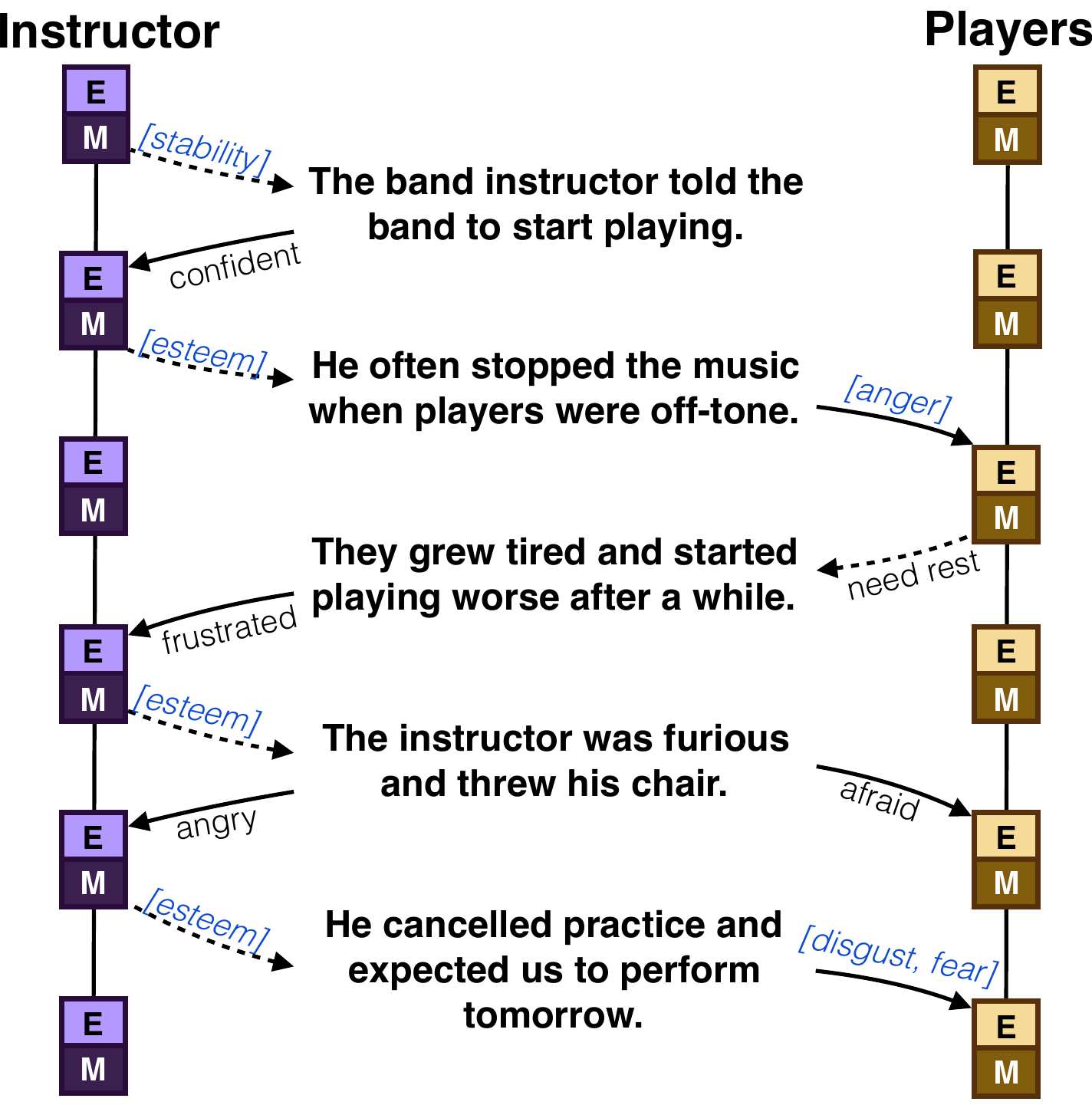}
\vspace*{-3mm}
\caption{ A story example with partial annotations for motivations (dashed) and emotional  reactions (solid). 
Open text explanations are in black (e.g., ``frustrated'') and formal theory labels are in blue with brackets (e.g., ``[esteem]'').}
\vspace*{-2mm}
\label{introfigure}
\end{centering}
\end{figure}

\begin{figure*}[t]
    \centering
    \hspace{-25pt}
    \includegraphics[width=.9\textwidth]{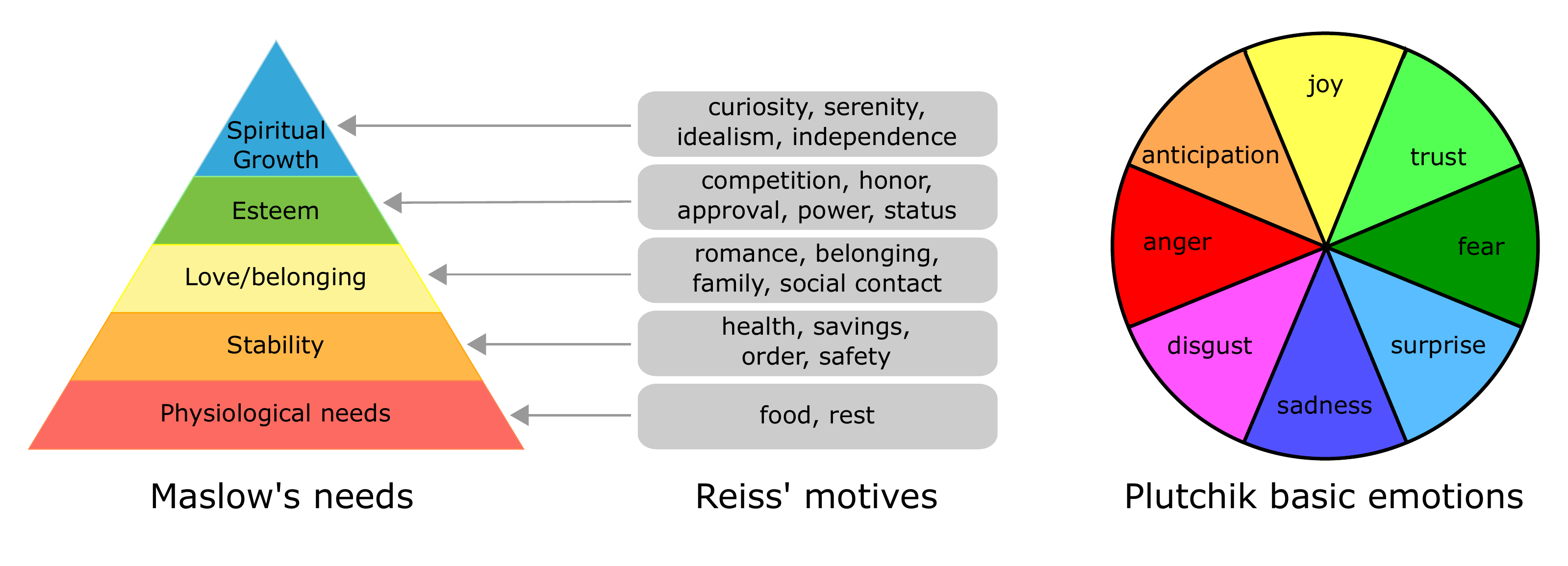}
    \vspace*{-6mm}
    \caption{Theories of Motivation (Maslow and Reiss) and Emotional Reaction (Plutchik).}
    \label{fig:annot_categories}
\end{figure*}

In this paper, we construct a new annotation formalism to densely label commonsense short stories \citep{Mostafazadeh2016-ei} in terms of 
the mental states of the characters. 
The resulting dataset offers three unique properties. 
First, as highlighted in Figure~\ref{introfigure}, the dataset provides a fully-specified chain of \emph{motivations} and \emph{emotional reactions} for each story character as pre- and post-conditions of events. Second, the annotations include state changes for entities even when they are not mentioned directly in a sentence (e.g., in the fourth sentence in Figure~\ref{introfigure},  
players would feel \emph{afraid} as a result of the instructor throwing a chair), 
thereby capturing implied effects unstated in the story. Finally, the annotations encompass both formal labels from multiple theories of psychology \cite{Maslow1943rl,Reiss2004ex,Plutchik1980jg} as well as open text descriptions of motivations and emotions, providing a comprehensive mapping between open text explanations and label categories (e.g., ``to spend time with her son'' $\rightarrow$ Maslow's category \textit{love}).  
Our corpus\footnote{We make our dataset publicly available at \url{https://uwnlp.github.io/storycommonsense/}} spans across 15k stories, amounting to 300k low-level annotations for around 150k character-line pairs.  

Using our new corpus, we present baseline performance on two new tasks focusing on mental state tracking of story characters: \emph{categorizing} motivations and emotional reactions using theory labels, as well as \emph{describing} motivations and emotional reactions using open text.  Empirical results demonstrate that existing neural network models including those with explicit or latent entity representations achieve promising results.



\section{Mental State Representations}
Understanding people's actions, motivations, and emotions has been 
a recurring research focus across several disciplines including philosophy and psychology \cite{schachter1962cognitive,burke1969grammar,lazarus1991progress,goldman2015theory}. 
We draw from these prior works to derive a set of categorical labels for annotating the step-by-step causal dynamics between the mental states of story characters and the events they experience. 



\subsection{Motivation Theories}

We use two popular theories of motivation:  the ``hierarchy of needs'' of \citet{Maslow1943rl} and  the ``basic motives'' of \citet{Reiss2004ex} to compile 5 coarse-grained and 19 fine-grained motivation categories, shown in Figure~\ref{fig:annot_categories}.
%
Maslow's ``hierarchy of needs" are comprised of five categories, ranging from \textit{physiological needs} to \textit{spiritual growth}, which we use as coarse-level categories.
%
\citet{Reiss2004ex} proposes 19 more fine-grained categories that provide a more informative range of motivations. For example, even though they both relate to the \textit{physiological needs} Maslow category, the \textit{food} and \textit{rest} motives from \citet{Reiss2004ex} are very different. While the Reiss theory allows for finer-grained annotations of motivation, the larger set of abstract concepts can be overwhelming for annotators.  Motivated by \citet{reissMaslowWebsite}, we design a hybrid approach, where Reiss labels are annotated as sub-categories of Maslow categories.



\subsection{Emotion Theory} 
Among several theories of emotion, we work with the ``wheel of emotions'' of \citet{Plutchik1980jg}, as it has been a common choice in prior literature on emotion categorization \cite{MohammadEmoLex,Zhou2016Emotion}.  We use the eight basic emotional dimensions as illustrated in Figure~\ref{fig:annot_categories}.

\subsection{Mental State Explanations}
In addition to the motivation and emotion categories derived from psychology theories, we also obtain open text descriptions of character mental states. 
These open text descriptions 
allow learning computational models that can \emph{explain} the mental states of characters in natural language, which is likely to be more accessible and informative to end users than having theory categories alone. Collecting both theory categories and open text also allows us to learn the automatic mappings between the two, which  generalizes the previous work of \citet{MohammadEmoLex} on emotion category mappings. 



\begin{figure*}[tb]
    \centering
    \includegraphics[width=.85\textwidth]{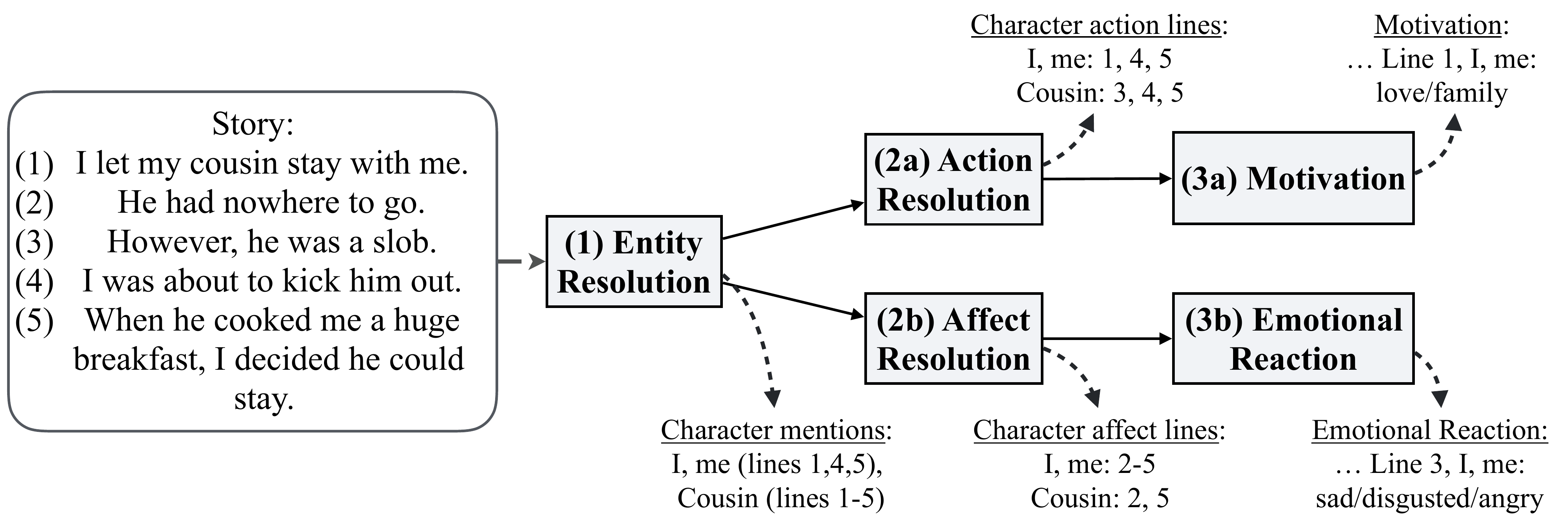}
    \vspace*{-3mm}
    \caption{The annotation pipeline for the fine-grained annotations with an example story.}
    \vspace*{-1mm}
    \label{pipeline:diagram}
\end{figure*}

\section{Annotation Framework}
\label{sec:data}
%
In this study, we choose to annotate the simple 
commonsense stories introduced by \citet{Mostafazadeh2016-ei}. Despite their simplicity, these stories pose a significant challenge to natural language understanding models \cite{SharedTask}. In addition, they depict multiple interactions between story characters, presenting rich opportunities to reason about character motivations and reactions. Furthermore, there are more than 98k such stories currently available covering a wide range of everyday scenarios.
\vspace{-1mm}
\paragraph{Unique Challenges} 
While there have been a variety of annotated resources developed on the related topics of sentiment analysis \cite{MohammadEmoLex,deng2015mpqa}, entity tracking \cite{Hoffart2011RobustDO,babi}, and story understanding \cite{Goyal2010AutomaticallyPP,Ouyang2015ModelingRE,Lukin2016PersonaBankAC}, our study is the first to annotate the full chains of mental state effects for story characters.  This poses several unique challenges as annotations require (1) interpreting discourse 
(2) 
understanding implicit causal effects, and (3) understanding formal psychology theory categories. 
In prior literature, annotations of this complexity have typically been performed by experts \cite{deng2015mpqa,Ouyang2015ModelingRE}. While reliable, these annotations are prohibitively expensive to scale up. Therefore, we introduce 
a new annotation framework that pipelines a set of smaller isolated tasks as illustrated in Figure~\ref{pipeline:diagram}.  All annotations were collected using crowdsourced workers from Amazon Mechanical Turk.
\begin{figure*}[tb]
  \centering
  \includegraphics[width=\textwidth]{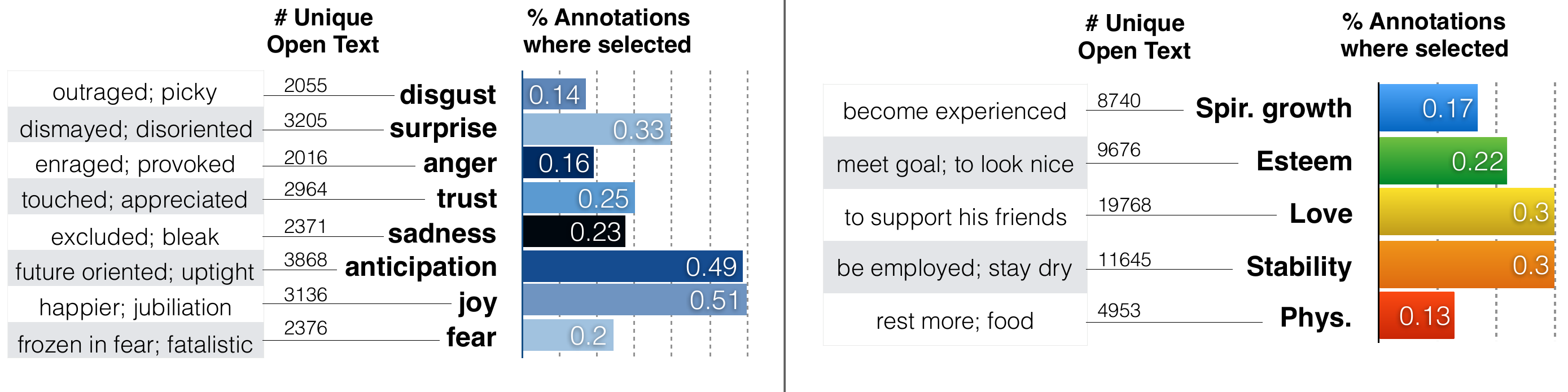}
 \vspace*{-8mm}
\caption{Examples of open-text explanations that annotators provided corresponding with the categories they selected.  The bars on the right of the categories represent the percentage of lines where annotators selected that category (out of those character-line pairs with positive motivation/emotional reaction).}
\label{lbldistrib}
\end{figure*}

\subsection{Annotation Pipeline}


We describe the components and workflow of the full annotation pipeline shown in Figure~\ref{pipeline:diagram} below. The example story in the figure is used to illustrate the output of various steps in the pipeline (full annotations for this example are in the appendix). 


\paragraph{(1) Entity Resolution}
The first task in the pipeline aims to discover (1) the set of characters $E_i$ in each story $i$ and (2) the set of sentences $S_{ij}$ in which a specific character $j \in E_i$ is explicitly mentioned. For example, in the story in Figure~\ref{pipeline:diagram}, the characters identified by annotators are ``I/me'' and ``My cousin'', whom appear in sentences $\{ 1,4,5 \}$ and $\{1,2,3,4,5\}$, respectively.
%

We use $S_{ij}$ to control the workflow of later parts of the pipeline 
by pruning future tasks for sentences that are not tied to characters. Because $S_{ij}$ is used to prune follow-up tasks, we take a high recall strategy to include all sentences that at least one annotator selected.



\paragraph{(2a) Action Resolution} 
The next task identifies whether a character $j$ appearing in a sentence $k$ is taking any action to which a motivation can be attributed. We perform action resolution only for sentences $k \in S_{ij}$.
In the running example, we would want to know that the cousin in line 2 is not doing anything intentional, allowing us to omit this line in the next pipeline stage (3a) where a character's motives are annotated.
Description of state (e.g., ``Alex is feeling blue'') or passive event participation (e.g., ``Alex trips'') are not considered volitional acts for which the character may have an underlying motive. 
%
%
%
For each line and story character pair, we obtain 4 annotations. Because pairs can still be filtered out in the next stage of annotation, we select a generous threshold where only 2 annotators must vote that an intentional action took place for the sentence to be used as an input to the motivation annotation task (3a).

\paragraph{(2b) Affect Resolution} 
This task aims to identify all of the lines where a story character $j$ has an emotional reaction. 
Importantly, it is often possible to infer the emotional reaction of a character $j$ even when the character does not explicitly appear in a sentence $k$. 
For instance, in Figure~\ref{pipeline:diagram}, we want to annotate the narrator's reaction to line 2 even though they are not mentioned because their emotional response is inferrable.
We obtain 4 annotations per character per line.  The lines with at least 2 annotators voting are used as input for the next task: (3b) emotional reaction.

\paragraph{(3a) Motivation}
We use the output from the action resolution stage (2a) to ask workers to annotate character motives in lines where they intentionally initiate an event.  We provide 3 annotators a line from a story, the preceding lines, and a specific character.  They are asked to produce a free response sentence describing what causes the character's behavior in that line and to select the most related Maslow categories and Reiss subcategories. In Figure~\ref{pipeline:diagram}, an annotator described the motivation of the narrator in line 1 as wanting ``to have company'' and then selected the \textit{love} (Maslow) and \textit{family} (Reiss) as categorical labels. Because many annotators are not familiar with motivational theories, we require them to complete a tutorial the first time they attempt the task.  

\paragraph{(3b) Emotional Reaction}
Simultaneously, we use the output from the affect resolution stage (2b) to ask workers what the emotional response of a character is immediately following a line in which they are affected.  As with the motives, we give 3 annotators a line from a story, its previous context, and a specific character.  We ask them to describe in open text how the character will feel following the event in the sentence (up to three emotions).  As a follow-up, we ask workers to compare their free responses against Plutchik categories by using 3-point likert ratings. In Figure~\ref{pipeline:diagram}, we include a response for the emotional reaction of the narrator in line 1.  Even though the narrator was not mentioned directly in that line, an annotator recorded that they will react to their cousin being a slob by feeling ``annoyed'' and selected the Plutchik categories for \textit{sadness}, \textit{disgust} and \textit{anger}.

\begin{table}[tb]
\centering
\begin{tabular}{lrrr}
 & & \multicolumn{2}{c}{Fine-grained}\\
 \cmidrule{3-4}
& train & dev & test \\
\# annotated stories         & 10000  &  2500   &  2500\\
\# characters / story        & 2.03  &  2.02   &  1.82\\
\# char-lines  w/ motiv     &  40154  &  8762  &  6831\\
\# char-lines  w/ emot      &  76613  &  14532 &  13785
\end{tabular}
\caption{Annotated data statistics for each dataset
}
\label{datastats}
\end{table}
\vspace*{-1mm}
\subsection{Dataset Statistics and Insights}

\begin{figure*}[tb]
\begin{centering}
\includegraphics[width=.95\textwidth, page=1]{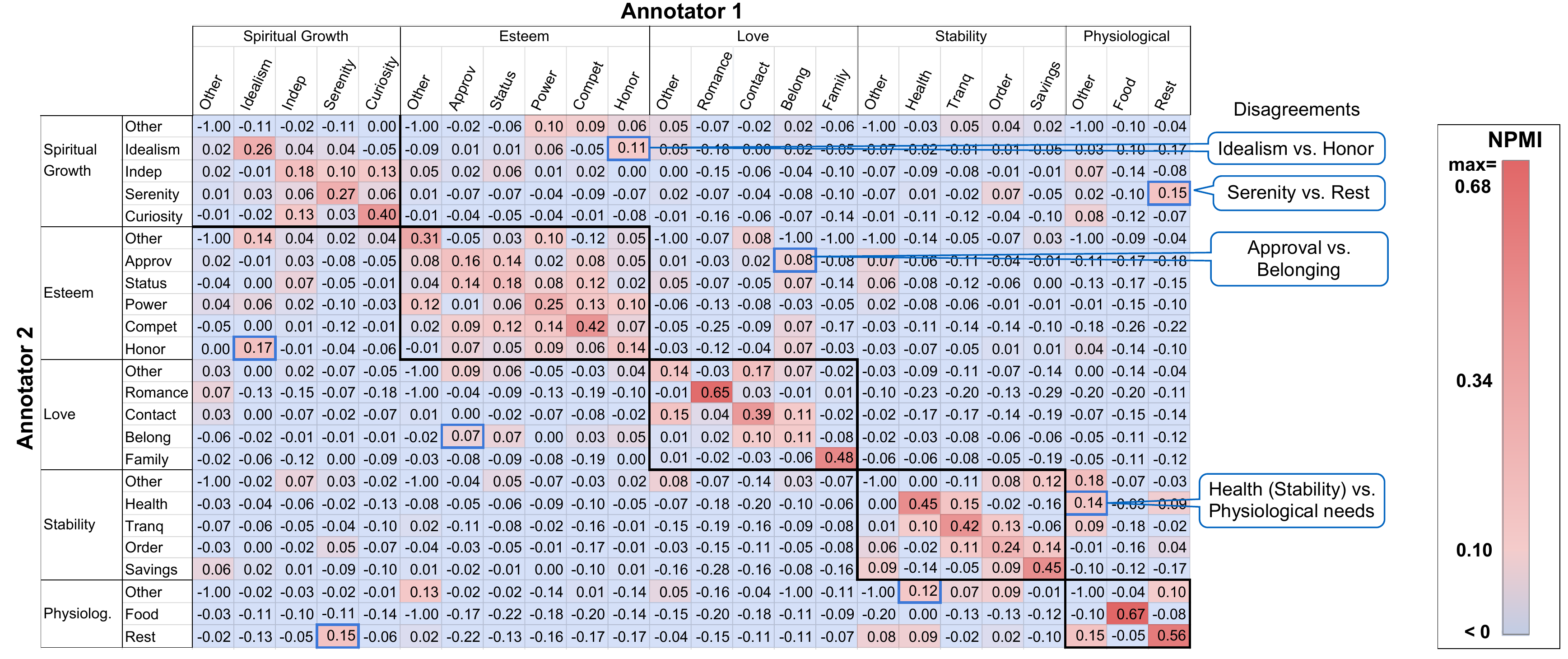}
\vspace*{-2mm}
\caption{NPMI confusion matrix on motivational categories for all annotator pairs with color scaling for legibility. The highest values are generally along diagonal or within Maslow categories (outlined in black).  We highlight a few common points of disagreement between thematically similar categories.}
\label{maslow-heatmap}
\end{centering}
\end{figure*}
\paragraph{Cost}
The tasks corresponding to the theory category assignments are the hardest and most expensive in the pipeline ($\sim$\$4 per story). 
Therefore, we obtain theory category labels only for a third of our annotated stories, which we assign to the development and test sets. The training data is annotated with a shortened pipeline with only open text descriptions of motivations and emotional reactions from two workers ($\sim$\$1 per story). 

\paragraph{Scale} 
Our dataset to date includes a total of 300k low-level annotations for motivation and emotion across 15,000 stories (randomly selected from the ROC story training set).  It covers over 150,000 character-line pairs, in which 56k character-line pairs have an annotated motivation and 105k have an annotated change in emotion (i.e. a label other than \texttt{none}). Table ~\ref{datastats} shows the break down across training, development, and test splits. Figure~\ref{lbldistrib} shows the frequency of different labels being selected for motivational and emotional categories in cases with positive change. 
%


\begin{table}[tb]
\centering
\begin{tabular}{@{}clrrr@{}}
\toprule
\textbf{Label Type} & \textbf{} & \textbf{\begin{tabular}[c]{@{}r@{}}PPA\end{tabular}} & \textbf{\begin{tabular}[c]{@{}r@{}}KA\end{tabular}} & \textbf{\begin{tabular}[c]{@{}r@{}}\%  Agree w/\\ Maj. Lbl\end{tabular}} \\ \midrule
\multirow{2}{*}{Maslow} & Dev & .77 & .30 & 0.88 \\
 & Test & .77 & .31 & 0.89 \\ \midrule
\multirow{2}{*}{Reiss} & Dev & .91 & .24 &  0.95 \\
 & Test & .91 & .24 &  0.95 \\ \midrule
\multirow{2}{*}{\begin{tabular}[c]{@{}c@{}}Plutchik\end{tabular}} & Dev & .71 & .32 & 0.84 \\
 & Test & .70 & .29 & 0.83 \\ \bottomrule
\end{tabular}
\caption{Agreement Statistics (PPA = Pairwise percent agreement of worker responses per binary category, KA= Krippendorff's Alpha)}
\label{kappastats}
\end{table}

\vspace*{-1mm}
\paragraph{Agreements} 
For quality control, we 
removed workers who consistently produced low-quality work, as discussed in the Appendix.
In the categorization sets (Maslow, Reiss and Plutchik), we compare the performance of annotators by treating each individual category as a binary label (1 if they included the category in their set of responses) and averaging the agreement per category.  For Plutchik scores, we count `moderately associated' ratings as agreeing with `highly' associated' ratings. 
The percent agreement and Krippendorff's alpha 
are shown in Table~\ref{kappastats}.  We also compute the percent agreement between the individual annotations and the majority labels.
\footnote{Majority label for the motivation categories is what was agreed upon by at least two annotators per category.  For emotion categories, we averaged the point-wise ratings and counted a category if the average rating was $\geq 2$.}

These scores are difficult to interpret by themselves, however, as annotator agreement in our categorization system has a number of properties that are not accounted for by these metrics  (disagreement preferences -- joy and trust are closer than joy and anger -- that are difficult to quantify in a principled way, hierarchical categories mapping Reiss subcategories from Maslow categories,  skewed category distributions that inflate PPA and deflate KA scores, and annotators that could select multiple labels for the same examples).

To provide a clearer understanding of agreement within this dataset, we create aggregated confusion matrices for annotator pairs.  First, we sum the counts of combinations of answers between all paired annotations (excluding \texttt{none} labels).  If an annotator selected multiple categories, we split the count uniformly among the selected categories.  We compute NPMI over the total confusion matrix.  In Figure~\ref{maslow-heatmap}, we show the NPMI confusion matrix for motivational categories.  

In the motivation annotations, we find the highest scores on the diagonal (i.e., Reiss agreement), with most confusions occurring between Reiss motives in the same Maslow category (outlined black in Figure~\ref{maslow-heatmap}).  Other disagreements generally involve Reiss subcategories that are thematically similar, such as \textit{serenity} (mental relaxation) and \textit{rest} (physical relaxation).  We provide this analysis for Plutchik categories in the appendix, finding high scores along the diagonal with disagreements typically occurring between categories in a ``positive emotion'' cluster (\textit{joy}, \textit{trust}) or a ``negative emotion'' cluster (\textit{anger}, \textit{disgust},\textit{sadness}).

\section{Tasks}

\begin{figure}
\begin{centering}
\includegraphics[width=.95\columnwidth, page=2]{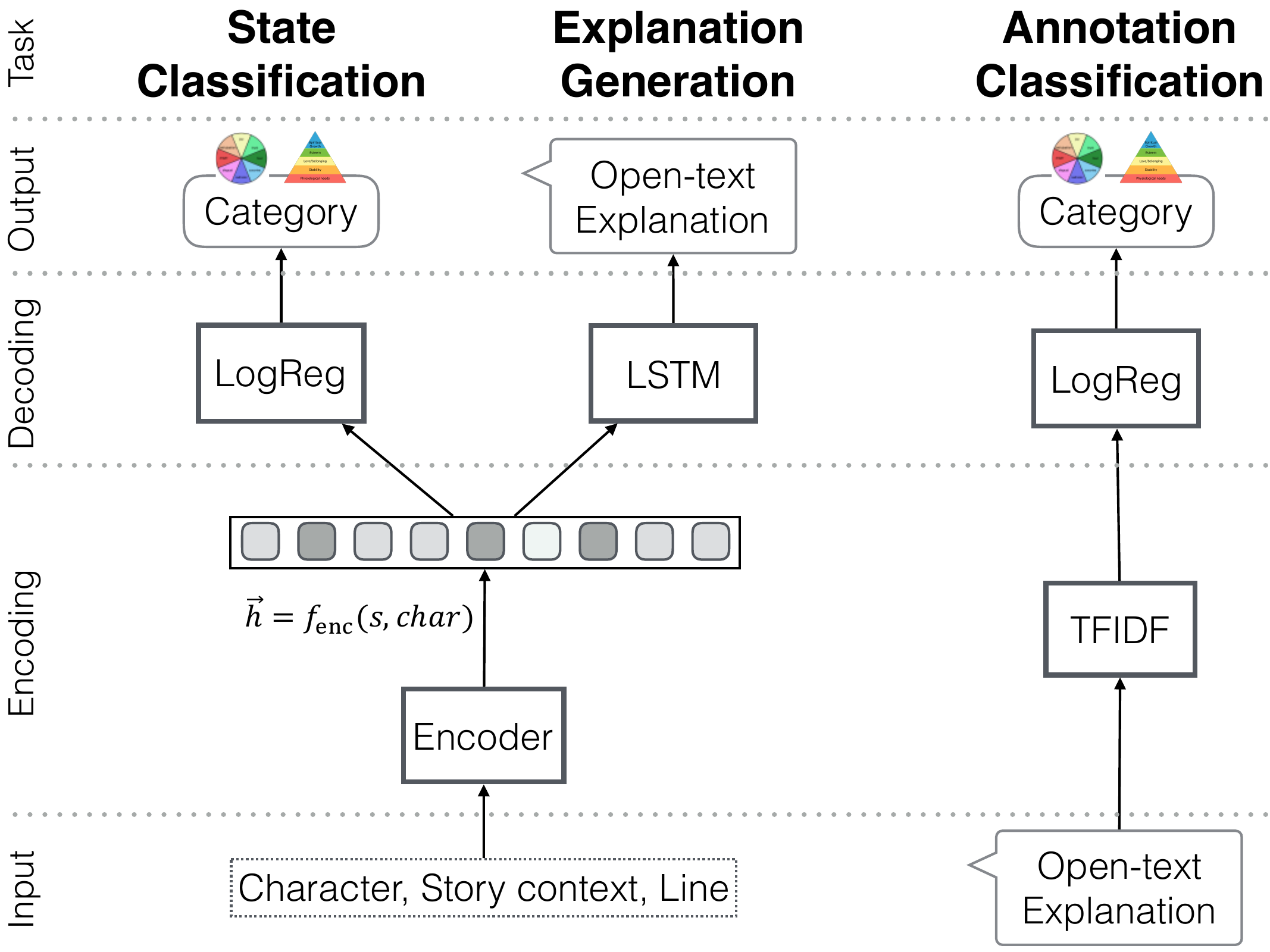}
\caption{General model architectures for three new task types}
\label{model-architecture}
\end{centering}
\end{figure}

The multiple modes covered by the annotations in this new dataset allow for multiple new tasks to be explored. We outline three task types below, covering a total of eight tasks on which to evaluate.  Differences between task type inputs and outputs are summarized in Figure~\ref{model-architecture}.

\paragraph{State Classification} The three primary tasks involve categorizing the psychological states of story characters for each of the label sets (Maslow, Reiss, Plutchik) collected for the dev and test splits of our dataset. In each classification task, a model is given a line of the story (along with optional preceding context lines) and a character and predicts the motivation (or emotional reaction). A binary label is predicted for each of the Maslow needs, Reiss motives or Plutchik categories.
\vspace{-1mm}
\paragraph{Annotation Classification} Because the dev and test sets contain paired classification labels and free text explanations, we propose three tasks where a model must predict the correct Maslow/Reiss/Plutchik label given an emotional reaction or motivation explanation.
\vspace{-1mm}
\paragraph{Explanation Generation} Finally, we can use the free text explanations to train models to describe the psychological state of a character in free text (examples in Figure~\ref{lbldistrib}). These explanations allow for two conditional generation tasks where the model must generate the words describing the emotional reaction or motivation of the character.

\section{Baseline Models}
The general model architectures for the three tasks are shown in Figure~\ref{model-architecture}.  We describe each model component below.  The state classification and explanation generation models could be trained separately or in a multi-task set-up.

In the state classification and explanation generation tasks, a model is given a line from a story $\mathbf{x}^s$ containing $N$ words $\{w^s_0, w^s_1, \dots, w^s_N\}$ from vocabulary $V$, a character in that story $e_j \in E$ where $E$ is the set of characters in the story, and (optionally) the preceding sentences in the story $\mathbf{C} = \{\mathbf{x}^0 \dots, \mathbf{x}^{s-1}\}$ containing words from vocabulary $V$. A representation for a character's psychological state is encoded as:
\vspace*{-1mm}
\begin{equation}
    \mathbf{h}^e = \mathrm{Encoder}(\mathbf{x}^s, \mathbf{C}[e_j])
\end{equation}

\noindent where $\mathbf{C}[e_j]$ corresponds to the concatenated subset of sentences in $\mathbf{C}$ where $e_j$ appears.

\subsection{Encoders}
\label{ssec:model} 

While the end classifier or decoder is different for each task, we use the same set of encoders based on word embeddings, common neural network architectures, or memory networks to formulate a representation of the sentence and character, $\mathbf{h}^e$. Unless specified, $\mathbf{h}^e$ is computed by encoding separate vector representations for the sentence ($\mathbf{x}^s \rightarrow \mathbf{h}^s$) and character-specific context ($\mathbf{C}[e_j] \rightarrow \mathbf{h}^c$) and concatenating these encodings ($\mathbf{h}^e = [\mathbf{h}^c; \mathbf{h}^s]$). 
We describe the encoders below:

\paragraph{TF-IDF} We learn a TD-IDF model on the full training corpus of \citet{Mostafazadeh2016-ei} (excluding the stories in our dev/test sets).
To encode the sentence, we extract TF-IDF features for its words, yielding $v^s \in \mathcal{R}^V$. A projection and non-linearity is applied to these features to yield $\mathbf{h}^s$:
\vspace*{-1mm}
\begin{equation}
    \mathbf{h}^s = \phi(W_s v^s + b_s) \label{eq:embproj}
\end{equation}

\noindent where $W_s \in \mathcal{R}^{d \times H}$. The character vector $\mathbf{h}^c$ is encoded in the same way on sentences in the context pertaining to the character. 

\paragraph{GloVe} We extract pretrained Glove vectors \cite{pennington2014glove} for each word in $V$. The word embeddings are max-pooled, yielding embedding $v^s \in \mathcal{R}^H$, where $H$ is the dimensionality of the Glove vectors. Using this max-pooled representation, $\mathbf{h}^s$ and $\mathbf{h}^c$ are extracted in the same manner as for TF-IDF features (Equation~\ref{eq:embproj}).

\paragraph{CNN} We implement a CNN text categorization model using the same configuration as \citet{Kim2014ConvolutionalNN} to encode the sentence words. A sentence is represented as a matrix, $v^s \in \mathcal{R}^{M \times d}$ where each row is a word embedding $x^s_n$ for a word $w^s_n \in \mathbf{x}^s$.
\vspace{-1mm}
\begin{align}
    v^s &= [x^s_0, x^s_1, \dots, x^s_N] \\
    \mathbf{h}^s &= \mathrm{CNN}(v^s)
\end{align}

\noindent where CNN represents the categorization model from \citep{Kim2014ConvolutionalNN}. The character vector $\mathbf{h}^c$ is encoded in the same way with a separate CNN. Implementation details are provided in the appendix.

\paragraph{LSTM} 
A two-layer bi-LSTM encodes the sentence words and concatenates the final time step hidden states from both directions to yield $\mathbf{h}^s$. The character vector $\mathbf{h}^c$ is encoded the same way.



\paragraph{REN} We use the ``tied" recurrent entity network from \citet{ren}. A memory cell $m$ is initialized for each of the $J$ characters in the story, $E = \{e_0,\dots, e_J\}$. The REN reads documents one sentence at a time and updates $m_j$ for $e_j \in E$ after reading each sentence. Unlike the previous encoders, all sentences of the context $\mathbf{C}$ are given to the REN along with the sentence $\mathbf{x}^s$. The model learns to distribute encoded information to the correct memory cells. The representation passed to the downstream model is:
\vspace*{-1mm}
\begin{equation}
    \mathbf{h}^e = \{m_j\}^s
\end{equation}

\noindent where $\{m_j\}^s$ is the memory vector in the cell corresponding to $e_j$ after reading $\mathbf{x}^s$. Implementation details are provided in the appendix. 

\paragraph{NPN} We also include the neural process network from \citet{Bosselut17} with ``tied" entities, but ``untied" actions that are not grounded to particular concepts. The memory is initialized and accessed similarly as the REN. Exact implementation details are provided in the appendix.

\subsection{State Classifier}
\label{ssec:model:class}
Once the sentence-character encoding $\mathbf{h}^e$ is extracted, the state classifier predicts a binary label $\hat y_z$ for every category $z \in \mathcal{Z}$ where $\mathcal{Z}$ is the set of category labels for a particular psychological theory (e.g., disgust, surprise, etc. in the Plutchik wheel). We use logistic regression as a classifier:
\vspace*{-1mm}
\begin{equation}
    \hat y_z = \sigma (W_z \mathbf{h}^e + b_z) \label{eq:class}
\end{equation}

\noindent where $W_z$ and $b_z$ are a label-specific set of weights and biases for classifying each label $z \in \mathcal{Z}$.

\subsection{Explanation Generator} The explanation generator is a single-layer LSTM \cite{hochreiter1997long} that receives the encoded sentence-character representation $\mathbf{h}^e$ and predicts each word $y_t$ in the explanation using the same method from \citet{seq2seq}. Implementation details are provided in the appendix.

\subsection{Annotation Classifier}

For annotation classification tasks, words from open-text explanations are encoded with TF-IDF features. The same classifier architecture from Section~\ref{ssec:model:class} is used to predict the labels.

\section{Experimental Setup}

\begin{table*}[t]
\centering
\resizebox{0.8\linewidth}{!}{
\begin{tabular}{l  rrr  rrr  rrr}
\toprule
 \multirow{2}{*}{Model}          & \multicolumn{3}{c}{Maslow} & \multicolumn{3}{c}{Reiss} & \multicolumn{3}{c}{Plutchik} \\
          & P     & R     & F1    & P     & R     & F1    & P     & R     & F1      \\
\midrule 
Random    &  7.45 & 49.99 & 12.96 &  1.76 & 50.02 &  3.40 & 10.35 & 50.00 & 17.15 \\
Random (Weighted)   & 8.10 & 8.89 & 8.48 &  2.25 & 2.40 &  2.32 & 12.28 & 11.79 & 12.03 \\
\midrule
TF-IDF &                30.10 & 21.21 & 24.88 & 18.40 & 20.67 & 19.46 &                     20.05 & 24.11 & 21.90 \\
+ Entity Context &      29.79 & 34.56 & 32.00 & 20.55 & 24.81 & 22.48 &                     22.71 & 25.24 & 23.91\\
\midrule
GloVe &                 25.15 & 29.70 & 27.24 & 16.65 & 18.83 & 17.67 &                     15.19 & 30.56 & 20.29 \\
+ Entity Context &      27.02 & 37.00 & 31.23 & 16.99 & 26.08 & 20.58 &                     19.47 & \textbf{46.65} & 27.48\\
\midrule
LSTM & 24.64 & 35.30 & 29.02 & 19.91 & 19.76 & 19.84 &                                      20.27 & 30.37 & 24.31 \\
+ Entity Context & \textbf{31.29} & 33.85 & 32.52 & 18.35 & 27.61 & 22.05 &                 23.98 & 31.41 & 27.20 \\
+ Explanation Training & 30.34 & 40.12 & 34.55 & \textbf{21.38} & 28.70 & \textbf{24.51} &  25.31 & 33.44 & 28.81 \\
\midrule
CNN \cite{Kim2014ConvolutionalNN} & 26.21 & 31.09 & 28.44 & 20.30 & 23.24 & 21.67 &         21.15 & 23.36 & 22.20 \\
+ Entity Context & 27.47 & 41.01 & 32.09 & 18.89 & 31.22 & 23.54 &                          24.32 & 30.76 & 27.16\\
+ Explanation Training & 29.30 & 44.18 & \textbf{35.23} & 17.87 & \textbf{37.52} & 24.21 &  24.47 & 38.87 & 30.04 \\
\midrule
REN \cite{ren}& 26.24 & 42.14 & 32.34 & 16.79 & 22.20 & 19.12 &                                       \textbf{26.22} & 33.26 & 29.32 \\
+ Explanation Training & 26.85 & \textbf{44.78} & 33.57 & 16.73 & 26.55 & 20.53 &           25.30 & 37.30 & 30.15 \\
\midrule
NPN \cite{Bosselut17} &                               24.27 & 44.16 & 31.33 & 13.13 & 26.44 & 17.55 & 21.98 & 37.31 & 27.66 \\
+ Explanation Training &            26.60 & 39.17 & 31.69 & 15.75 & 20.34 & 17.75 & 24.33 & 40.10 & \textbf{30.29} \\
\bottomrule
\end{tabular}}
\caption{State Classification Results}
\label{tab:track}
\end{table*}

\subsection{Training}

\paragraph{State Classification} The dev set $D$ is split into two portions of 80\% ($D_1$) and 20\% ($D_2)$. $D_1$ is used to train the classifier and encoder. $D_2$ is used to tune hyperparameters. The model is trained to minimize the weighted binary cross entropy of predicting a class label $y_z$ for each class $z$:
\vspace*{-1mm}
\begin{equation}
    \mathcal{L} = \sum_{z=1}^Z \gamma_z y_z \log \hat y_z + (1 - \gamma_z) (1 - y_z) \log (1 - \hat y_z) \label{eq:bce}
\end{equation}

\noindent where $Z$ is the number of labels in each of the three classifications tasks and $\gamma_z$ is defined as:

\begin{equation}
    \gamma_z = 1 - e^{-\sqrt{P(y_z)}}
\end{equation}

\noindent where $P(y_z)$ is the marginal class probability of a positive label for $z$ in the training set.
\vspace{-1mm}
\paragraph{Annotation Classification} The dev set is split in the same manner as for state classification. The TF-IDF features are trained on the set of training annotations $D_t$ coupled with those from $D_1$
. The model must minimize the same loss as in Equation~\ref{eq:bce}. Details are provided in the appendix.
\vspace{-1mm}
\paragraph{Explanation Generation}

We use the training set of open annotations to train a model to predict explanations. The decoder is trained to minimize the negative loglikelihood of predicting each word in the explanation of a character's state:
\vspace{-1mm}
\begin{equation}
L_{gen} = - \sum_{t=1}^{T} \log P(y_{t} | y_0, ..., y_{t-1}, \mathbf{h}^e) \label{eq:mle}
\end{equation}

\noindent where $\mathbf{h}^e$ is the sentence-character representation produced by an encoder from Section~\ref{ssec:model}.

\subsection{Metrics}

\paragraph{Classification} For the state and annotation classification task, we report the micro-averaged precision (P), recall (R), and F1 score of the Plutchik, Maslow, and Reiss prediction tasks. We report the results of selecting a label at random in the top two rows of Table~\ref{tab:track}. Note that random is low because the distribution of positive instances for each category is very uneven: macro-averaged positive class probabilities of 8.2, 1.7, and  9.9\% per category for Maslow, Reiss, and Plutchik respectively.
\vspace{-1mm}
\paragraph{Generation} Because explanations tend to be short sequences (Figure~\ref{lbldistrib}) with high levels of synonymy, traditional metrics such as BLEU are inadequate for evaluating generation quality. We use the vector average and vector extrema metrics from \citet{Liu2016HowNT} computed using the Glove vectors of generated and reference words. We report results in Table~\ref{tab:gen} on the dev set and compare to a baseline that randomly samples an example from the dev set as a generated sequence.

\subsection{Ablations}

\paragraph{Story Context vs. No Context} Our dataset is motivated by the importance of interpreting story context to categorize emotional reactions and motivations of characters. To test this importance, we ablate $\mathbf{h}^c$, the representation of the context sentences pertaining to the character, as an input to the state classifier for each encoder (except the REN and NPN). In Table~\ref{tab:track}, this ablation is the first row for each encoder presented.
\vspace{-1mm}
\paragraph{Explanation Pretraining}
Because the state classification and explanation generation tasks use the same models to encode the story, we explore initializing a classification encoder with parameters trained on the generation task. For the CNN, LSTM, and REN encoders, we pretrain a generator to produce emotion or motivation explanations. We use the parameters from the emotion or motivation explanation generators to initialize the Plutchik or Maslow/Reiss classifiers respectively.

\section{Experimental Results}
\label{sec:experiments}

\paragraph{State Classification} We show results on the test set for categorizing Maslow, Reiss, and Plutchik states in Table~\ref{tab:track}.  
Despite the difficulty of the task
, all models outperform the random baseline. Interestingly, the performance boost from adding entity-specific contextual information (i.e., not ablating $\mathbf{h}^c$) indicates that the models learn to condition on a character's previous experience to classify its mental state at the current time step.  
This effect can be seen in a story about a man whose flight is cancelled. The model without context predicts the same emotional reactions for the man, his wife and the pilot, but with context correctly predicts that the pilot 
will not have a reaction while predicting that the man and his wife will feel sad.


For the CNN, LSTM, REN, and NPN models, we also report results from pretraining encoder parameters using the free response annotations from the training set.  This pretraining offers a clear performance boost for all models on all three prediction tasks, showing that the parameters of the encoder can be pretrained on auxiliary tasks providing emotional and motivational state signal.

The best performing models in each task are most effective at predicting Maslow \textit{physiological} needs, 
Reiss \textit{food} motives
, and Plutchik reactions of \textit{joy}
. The relative ease of predicting motivations related to food (and physiological needs generally) may be because they involve a more limited and concrete set of actions such as eating or cooking.

\begin{table}[tb]
\centering
\begin{tabular}{l  c  c c}
\toprule
 & Maslow & Reiss & Plutchik \\
\midrule
TFIDF        &   64.81    &  48.60 &    53.44  \\   
\bottomrule
\end{tabular}
\caption{F1 scores of predicting correct category labels from free response annotations}
\label{tab:map}
\end{table}
\vspace{-1mm}
\paragraph{Annotation Classification}  Table~\ref{tab:map} shows that a simple model can learn to map open text responses to categorical labels. This further supports our hypothesis that pretraining a classification model on the free-response annotations could be helpful in boosting performance on the category prediction. 
\vspace{-1mm}
\paragraph{Explanation Generation} Finally, we provide results for the task of generating explanations of motivations and emotions in Table~\ref{tab:gen}.  Because the explanations are closely tied to emotional and motivation states, the randomly selected explanation can often be close in embedding space to the reference explanations, making the random baseline fairly competitive.  However, all models outperform the strong baseline on both metrics, indicating that the generated short explanations are closer semantically to the reference annotation. 

\begin{table}[t]
\centering
\begin{tabular}{l  c  c c c}
\toprule
\multirow{2}{*}{Model}          & \multicolumn{2}{c}{Motivation} & \multicolumn{2}{c}{Emotion} \\
& Avg & VE & Avg & VE \\
\midrule
Random    & 56.02     &  45.75  & 40.23    &   39.98        \\
\midrule
LSTM      & 58.48     & 51.07      &  52.47     & 52.30        \\
CNN       & 57.83     & 50.75      & 52.49    & 52.31      \\
REN       & \textbf{58.83}     & \textbf{51.79}      &  53.95    & 53.79       \\
NPN       &  57.77     & 51.77 & \textbf{54.02} &   \textbf{53.85}      \\

\bottomrule
\end{tabular}
\caption{Vector average and extrema scores for generation of annotation explanations}
\label{tab:gen}
\end{table}


\section{Related Work}

\paragraph{Mental State Annotations}

Incorporating emotion theories into NLP tasks has been explored in previous projects.
\citet{Ghosh2017-lj} modulate language model distributions by increasing the probability of words that express certain affective LIWC \cite{Tausczik2016-rd} categories.  More generally, various projects tackle the problem of generating text from a set of attributes like sentiment or generic-ness \cite{Ficler2017-ab,Dong2017-vr}.  Similarly, there is also a body of research in reasoning about commonsense stories and discourse \cite{Li2017-vw, Mostafazadeh2016-ei} or detecting emotional stimuli in stories \cite{Gui2017-op}.  Previous work in plot units \cite{Lehnert1981PlotUnits} developed formalisms for affect and mental state in story narratives that included motivations and reactions.
In our work, we collect mental state annotations for stories to used as a new resource in this space.

\vspace{-1mm}
\paragraph{Modeling Entity State}

Recently, novel works in language modeling \cite{Ji2017,reflm}, question answering \cite{ren}, and text generation \cite{Kiddon2016-ez,Bosselut17} have shown that modeling entity state explicitly can boost performance while providing a preliminary interface for interpreting a model's prediction. Entity modeling in these works, however, was limited to tracking entity reference \cite{Kiddon2016-ez,reflm,Ji2017}, recognizing entity state similarity \cite{ren} or predicting simple attributes from entity states \cite{Bosselut17}. Our work provides a new dataset for tracking emotional reactions and motivations of characters in stories.

\section{Conclusion}
We present a large scale dataset as a resource for training and evaluating mental state tracking of characters in short commonsense stories.  This dataset contains over 300k low-level annotations for character {\it motivations} and {\it emotional reactions}.  
We provide benchmark results on this new resource. Importantly, we show that modeling character-specific context and pretraining on free-response data can boost labeling performance.  
While our work only use information present in our dataset, we view our dataset as a future testbed for evaluating models trained on any number of resources for learning common sense about emotional reactions and motivations.


\section*{Acknowledgments}
%
We thank the reviewers for their insightful comments.
We also thank Bowen Wang, 
xlab members, Martha Palmer, Tim O'Gorman, Susan W. Brown, and Ghazaleh Kazeminejad 
for helpful discussions on inter-annotator agreement and the annotation pipeline. 
This work was supported in part by
NSF GRFP DGE-1256082, NSF IIS-1714566, IIS-1524371, Samsung AI, 
and DARPA CwC (W911NF-15-1-0543).

\bibliographystyle{acl_natbib}
\bibliography{references}
\clearpage
\appendix

\section{AMT Annotation Notes}
We ran AMT tasks in batches. We performed quality control tests after every 1-2 batches and banned workers who were consistently performing poorly (and removed their responses from our dataset).  For quality control, we used a combination of automatic methods (SGD over interannotator agreement using the quality control objectives described in the supplementary material of \citealp{yatskar2016}) and manual methods (e.g. searching for workers who always selected the same answers, who had strange answering patterns, who copied-and-pasted free responses) to identify workers with poor-quality work. Work by such annotators was reviewed, and these workers were banned if they were repeatedly not following instructions on a number of HIT's.

\begin{figure*}[tb]
\begin{centering}
\includegraphics[width=.95\textwidth, page=2,trim=0 2.cm 1.5cm 0, clip]{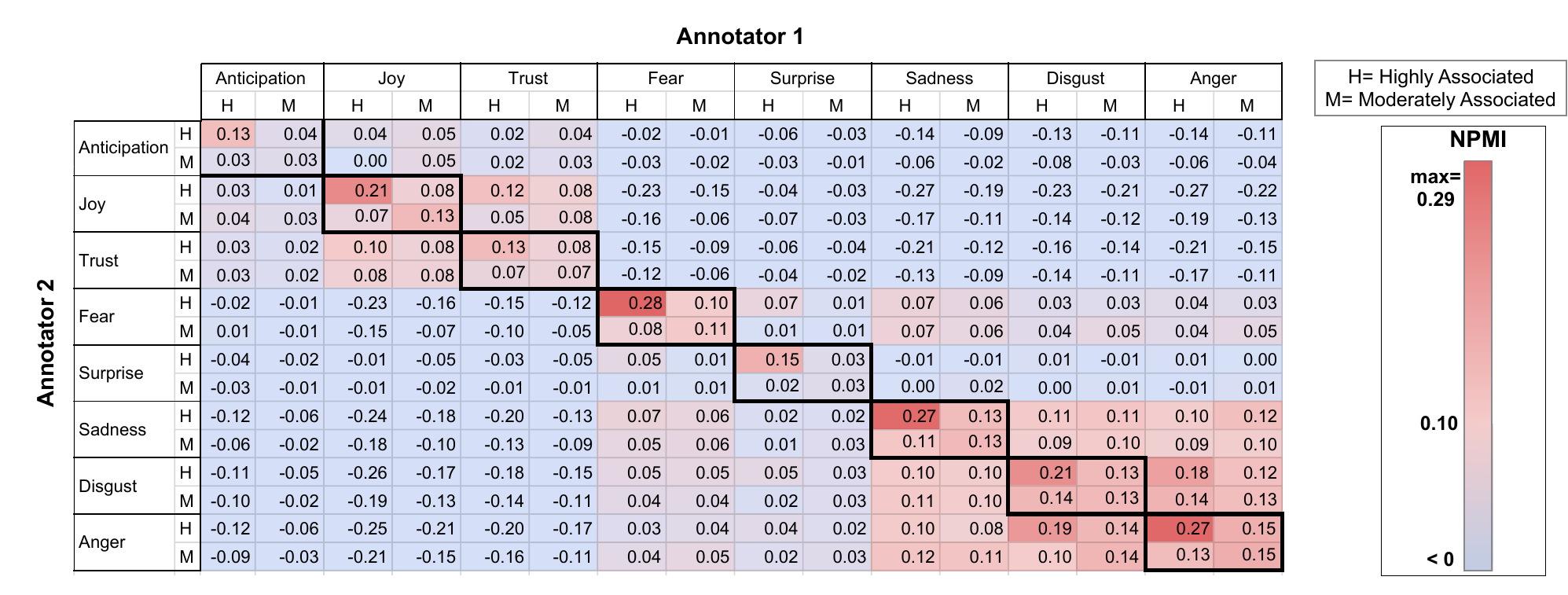}
\vspace*{-2mm}
\caption{NPMI confusion matrix on emotion categories for all annotator pairs with color scaling for legibility.}
\label{emotion-heatmap}
\end{centering}
\end{figure*}

\section{Emotion Agreement Matrix}

We include a NPMI confusion matrix for aggregated Plutchik paired responses in Figure~\ref{emotion-heatmap}.  Black boxes signify the same emotion but at different intensities (high vs. moderate).  In general higher co-occurring responses are along the diagonal.  However, we note that there are two main clusters coinciding with strongly positive emotions (joy and trust) and strongly negative emotions (anger, disgust, and sadness) where disagreements are more likely to occur.  To a lesser extent, there also is a slight co-occurrence between fear and the strongly negative emotions.

\section{Model Implementation Details}

All classification models are trained with the Adam Optimizer \cite{adam} with a learning rate 0.001 and gradient clipping if the norm of the gradients exceeds 1. We regularize with dropout layers whose probabilities are specific to each model. All models are trained with word embeddings of dimensionality 100 that are initialized with pretrained Glove vectors \cite{pennington2014glove}. For classification labels, we use the majority label among annotators for a particular character-line pair. 

\subsection{LSTM Classifier}

We train a 2-layer bidirectional LSTM encoder. The hidden states of the LSTM have dimensionality 100. We add dropout layers with p=0.5 in between the word embedding layer and the LSTM and between LSTM layers \cite{dropout}. We include a dropout layer with p=0.5 before the logistic regression classifier.

\subsection{CNN Classifier}

We follow the approach of \citet{Kim2014ConvolutionalNN} and train a CNN classifier with kernels of size 3, 4, and 5. We use 100 kernels of each size. We add a dropout layer with p=0.5 between the word embedding layer and the convolutional kernel layers. We include a dropout layer with p=0.5 before the logistic regression classifier. 

\subsection{REN Classifier}

We use the same implementation as \citet{ren} except that we remove the output module designed to encode questions and instead select the memory cell tied to the entity of interest for every training example. All equations from the input encoder and dynamic memory are identical to those of \cite{ren}. The input encoder computes a positionally weighted average of all the words in a sentence:

\begin{equation}
s_t = \sum_i f_i \odot e_i    
\end{equation}

\noindent where $e_i$ is a word embedding at index $i$ in a sentence, $f_i$ is a positional weight for that index in the sentence, and $s_t$ is a sentence representation. The dynamic memory is updated in the following way:

\begin{align}
    g_j &= \sigma(s_t^Th_j + s_t^Tw_j) \\
    \tilde h_j &= \phi(Uh_j + Vw_j + Ws_t) \\
    h_j &\leftarrow h_j + g_j \odot \tilde h_j \\
    h_j &\leftarrow \frac{h_j}{\vert\vert h_j \vert\vert}
\end{align}

\noindent where $h_j$ is the value stored in memory cell $j$, $w_j$ is a key corresponding to memory cell $j$, $U$, $V$, $W$ are projection matrices, and $\phi$ is a PReLU non-linearity. We initialize entity memory keys and entity memory values with the sum of the Glove vectors for all the words in the character name. Entity key values $w_j$ are locked during training. We use dropout with p=0.3 between the encoder and dynamic memory. 

\subsection{NPN Classifier} We use the same implementation as in \citet{Bosselut17} with a few modifications to the underlying architecture. First, we use the same encoder as for the REN \cite{ren}. We define a set of action function embeddings that can be applied to entities to change their state, $\mathbf{A}$. After each sentence, the model selects an action function embedding to use to change the state of the entity memory. Unlike in \citet{Bosselut17}, these action function embeddings are not tied to real actions and are instead treated as latent t  The dynamic memory is updated in the following way:

\begin{align}
    g_j &= \sigma(s_t^T W_1 [h_j, w_j]) \\
    \alpha_t &= softmax(MLP(s_t)) \\
    a_t &= \alpha_t ^T \mathbf{A} \\
    \tilde h_j &= \phi(W_3a_t + W_4s_t) \\
    h_j &\leftarrow (1 - g_j) h_j + g_j \odot \tilde h_j \\
    h_j &\leftarrow \frac{h_j}{\vert\vert h_j \vert\vert}
\end{align}

where $h_j$ is the value stored in memory cell $j$, $w_j$ is a key corresponding to memory cell $j$, $W_k$ are projection matrices, $\phi$ is a PReLU non-linearity and $\alpha_t$ is a distribution over possible action function embeddings. We initialize entity memory keys and entity memory values with the sum of the Glove vectors for all the words in the character name. Entity key values $w_j$ are locked during training. We use dropout with p=0.5 between the encoder and dynamic memory.

\subsection{LSTM Decoder}

For the explanation generation task, we train a single-layer LSTM with a learning rate of 0.0003 and gradient clipping when the norm of the gradients exceeds 1. When outputting words, we concatenate the original hidden state from the encoder $\mathbf{h}^e$ to the output of the LSTM decoder before predicting a word. Word embeddings are initialized with pretrained Glove vectors \cite{pennington2014glove}. In the generation task, the model must learn to generate individual annotations. 

\section{Experimental Details}
\subsection{Data used for Annotation Classification Task}

We split the development set into two parts, 80\% used for training ($D_1$), 20\% used for evaluating hyperparameters ($D_2$). We train a set of TF-IDF features for each word using all of the explanations from the real training set ($D_t$) and the portion of the development set used for training ($D_1$). We train a logistic regression classifier with L2 regularization. When training the classifier on $D_1$, we only train on examples where the explanation was written by a Turker who selected at least one Plutchik category label that was part of the majority set of Plutchik labels for the sentence the explanation and labels belong to. We prune $D_2$ and the test set similarly. We use individual annotations rather than majority labels to better learn specific fine-grained mappings.

\paragraph{Emotional Explanation to Plutchik Labels} We re-balance the dataset by sampling from the training set evenly among positive examples and negative examples for each category. We use L2 regularization with $\lambda = 0.1$ We include the full positive class distributions for each category in Table~\ref{tab:classdistrib}.

\paragraph{Motivation Explanation to Maslow Labels} We use L2 regularization with $\lambda = 0.01$. The dataset is not rebalanced.

\paragraph{Motivation Explanation to Reiss Labels} We use L2 regularization with $\lambda = 0.1$. The dataset is not rebalanced.

\begin{table*}[tb]
\centering
\begin{tabular}{ll|ll|llll}
Maslow &  & Plutchik &  & Reiss &  &  &  \\ \hline
Spiritual growth & 6.24 & Disgust & 5.06 & Status & 2.53 & Romance & 2.00 \\
Esteem & 8.41 & Surprise & 11.99 & Idealism & 0.55 & Savings & 2.41 \\
Love & 11.35 & Anger & 5.78 & Power & 1.01 & Contact & 3.81 \\
Stability & 10.46 & Trust & 9.14 & Family & 3.57 & Health & 1.74 \\
Physiological & 4.37 & Sadness & 7.82 & Food & 2.87 & Serenity & 0.58 \\
 &  & Anticipation & 16.65 & Independence & 1.29 & Curiosity & 2.58 \\
 &  & Joy & 15.8 & Belonging & 0.14 & Approval & 1.88 \\
 &  & Fear & 7.15 & Competition & 2.58 & Rest & 0.71 \\
 &  &  &  & Honor & 0.67 & Tranquility & 2.34 \\
 &  &  &  &  &  & Order & 2.56
\end{tabular}

\caption{Class Distribution (percent positive instances) per category.}
\label{tab:classdistrib}
\end{table*}

\begin{figure*}[tb]
    \centering
    \includegraphics[width=.99\textwidth]{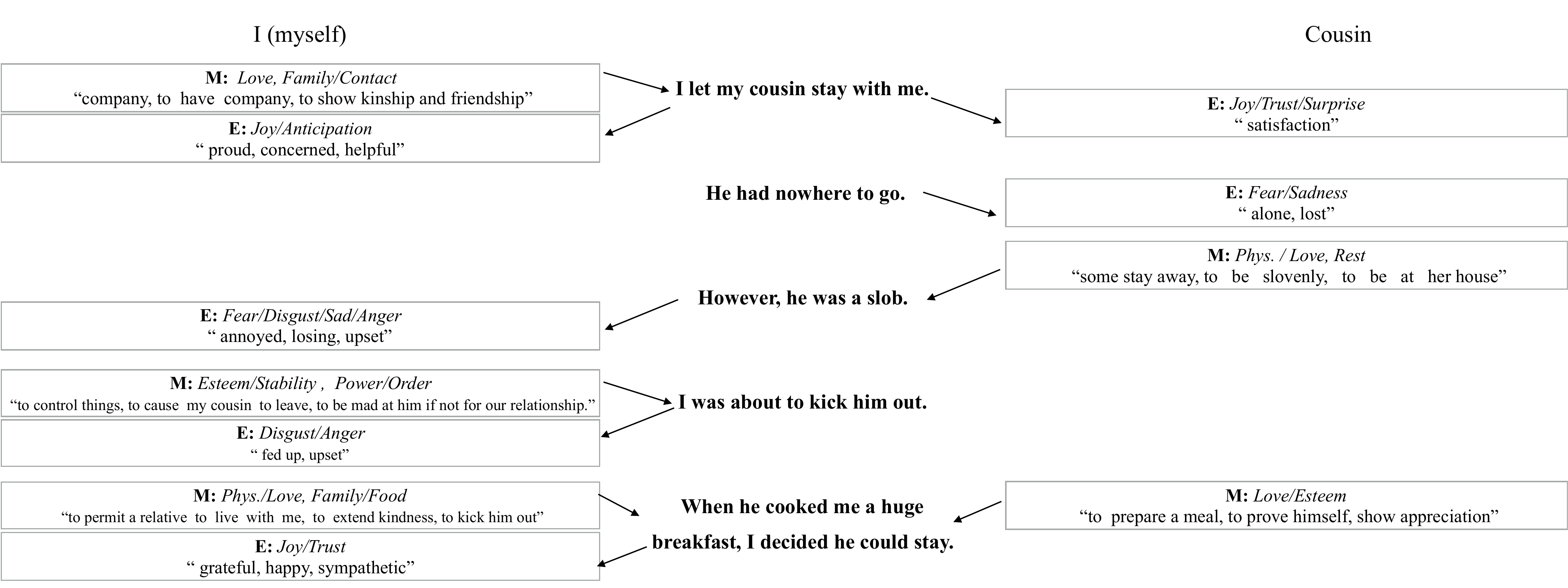}
    \caption{Fully annotated example from the annotation pipeline}
    \label{fig:appx:FullEx}
\end{figure*}

\end{document}